\pdfoutput=1

\documentclass[11pt]{article}

\usepackage{ACL2023}

\usepackage{times}
\usepackage{latexsym}

\usepackage[T1]{fontenc}

\usepackage[utf8]{inputenc}

\usepackage{microtype}

\usepackage{inconsolata}

\usepackage{graphicx}
\usepackage{booktabs}

\usepackage{color}

\definecolor{myred}{rgb}{1,0,0}
\definecolor{mygreen}{rgb}{0.40,0.64,0.25}

\usepackage{caption}
\usepackage{subcaption}

\usepackage{amsmath, bm, amsfonts}

\title{Few-Shot VQA with Frozen LLMs: A Tale of Two Approaches}

\author{Igor Sterner, Weizhe Lin, Jinghong Chen, Bill Byrne \\ Department of Engineering\\
University of Cambridge\\\texttt{\{is473, wl356, jc2124, wjb31\}@cam.ac.uk}}

\begin{document}
\maketitle
\begin{abstract}
    Two approaches have emerged to input images into large language models (LLMs).
    The first is to caption images into natural language.
    The second is to map image feature embeddings into the domain of the LLM and pass the mapped embeddings directly to the LLM.
    The majority of recent few-shot multimodal work reports performance using architectures that employ variations of one of these two approaches.
    But they overlook an important comparison between them.
    We design a controlled and focused experiment to compare these two approaches to few-shot visual question answering (VQA) with LLMs.
    Our findings indicate that for Flan-T5 XL, a 3B parameter LLM, connecting visual embeddings directly to the LLM embedding space does not guarantee improved performance over using image captions.
    In the zero-shot regime, we find using textual image captions is better.
    In the few-shot regimes, how the in-context examples are selected determines which is better.
\end{abstract}

\begin{figure}[b]
\footnotesize
    \centering
    \begin{minipage}{0.157\textwidth}
        \centering
        \includegraphics[height=.055\paperheight]{images/vqa_0.jpg}
        \par
        How many people are riding on the elephant?
        \textcolor{mygreen}{2}
    \end{minipage}
    \hfill
    \begin{minipage}{0.153\textwidth}
        \centering
        \includegraphics[height=.055\paperheight]{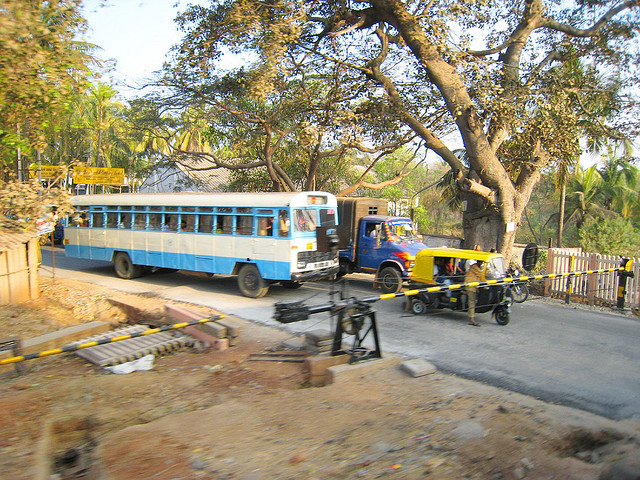}
        \par
        What color is the bus?
        \par
        \textcolor{mygreen}{blue and white}
    \end{minipage}
    \hfill
    \begin{minipage}{0.16\textwidth}
        \centering
        \includegraphics[height=.055\paperheight]{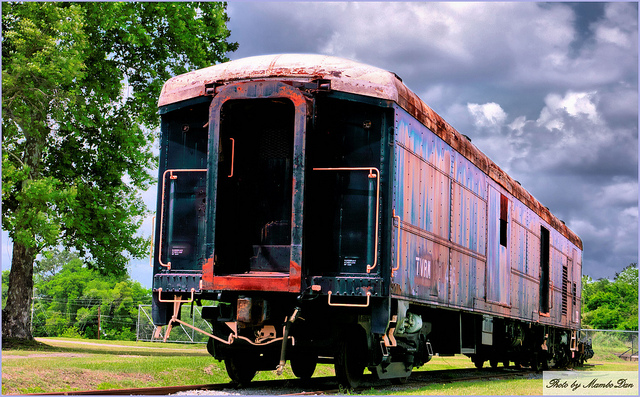}
        \par
        Does this train look to be in good working order?
        \textcolor{mygreen}{no}
    \end{minipage}
    \hfill
\caption{Examples of visual question answering. Source: \citet{goyal_making_2016}}
\label{fig:vqa-examples}
\end{figure}

\section{Introduction}

A fundamental feature of intelligence is the capability to integrate multiple modalities.
This enables many ambiguities to be reduced and supports knowledge acquisition.
In particular, a fundamental multimodal task is visual question answering (VQA).
VQA is the task of answering a question using information from an image.
Figure~\ref{fig:vqa-examples} shows three examples.

In the quest to integrate the modalities of vision and language, there has been a substantial effort to integrate the two dominant architectures in these fields: large language models (LLMs) and vision transformers \citep{alayrac_flamingo_2022, li_blip-2_2023, huang_language_2023, wang_simvlm_2021, chen_pali_2022}.
LLMs are a general-purpose interface for text-to-text tasks \citep{brown_language_2020}.
Vision transformers are able to transform images into a generic feature representation \citep{dosovitskiy_image_2021}. 
These models have, however, been trained separately.
This means that the representation of one can not be used in the other.

\begin{figure*}
    \centering
    \begin{subfigure}[b]{\textwidth}
        \centering  
        \includegraphics[scale=0.8]{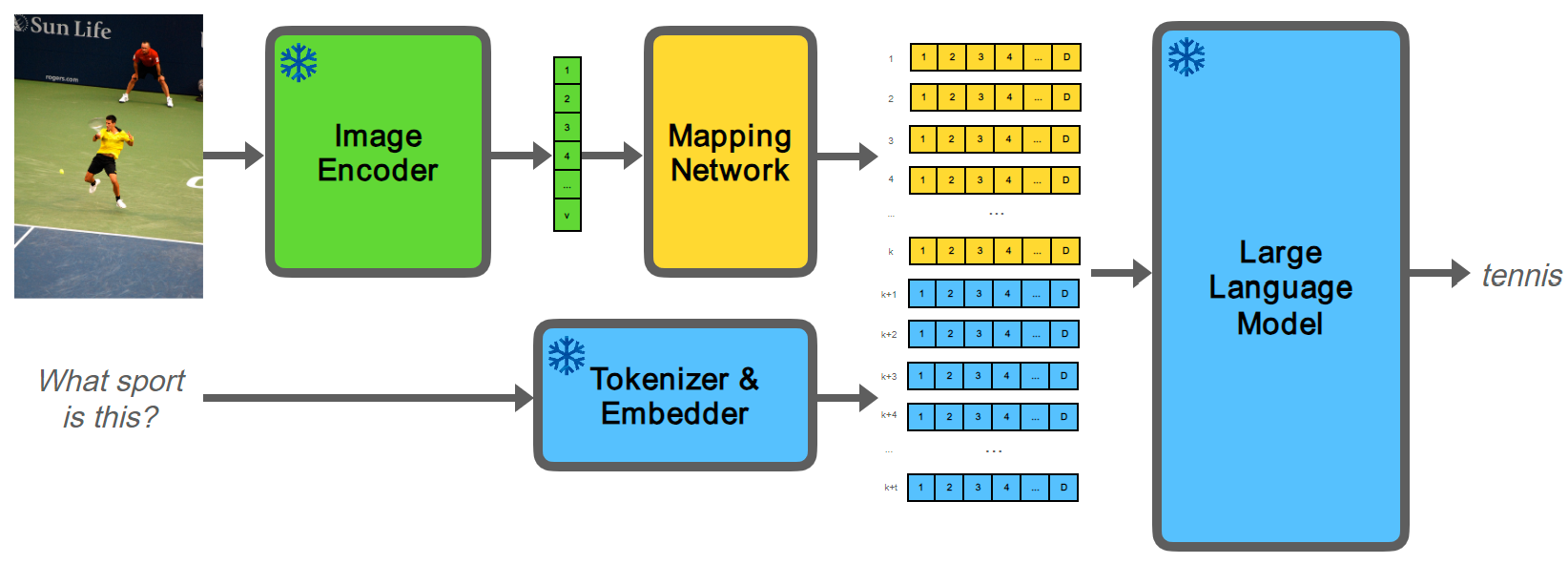}
        \caption{Embedding-based VQA}
        \label{fig:embed-vqa}
    \end{subfigure}
    \hfill
    \begin{subfigure}[b]{\textwidth}
        \centering
        \includegraphics[scale=0.8]{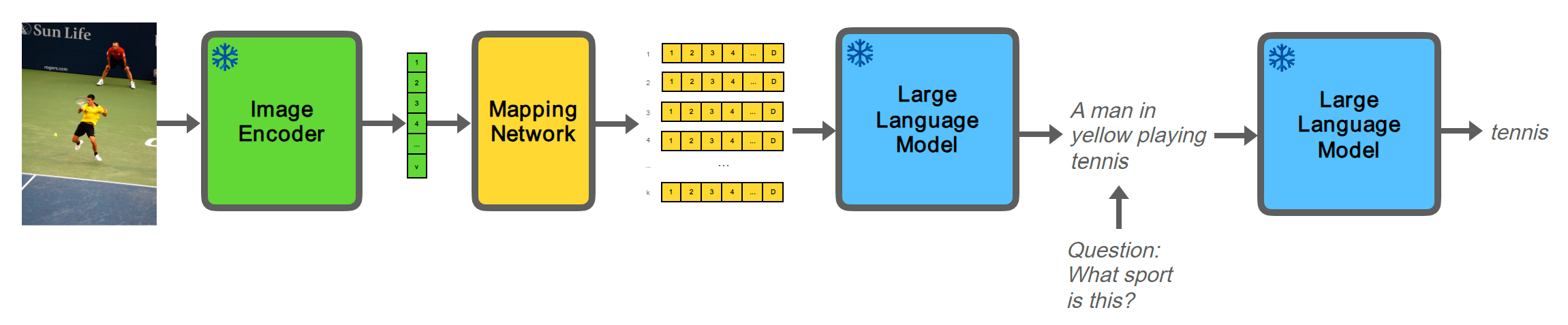}
        \caption{Caption-based VQA}
        \label{fig:caption-vqa}
    \end{subfigure}
    \caption{Two approaches to input images into LLMs for VQA. \includegraphics[width=.015\textwidth]{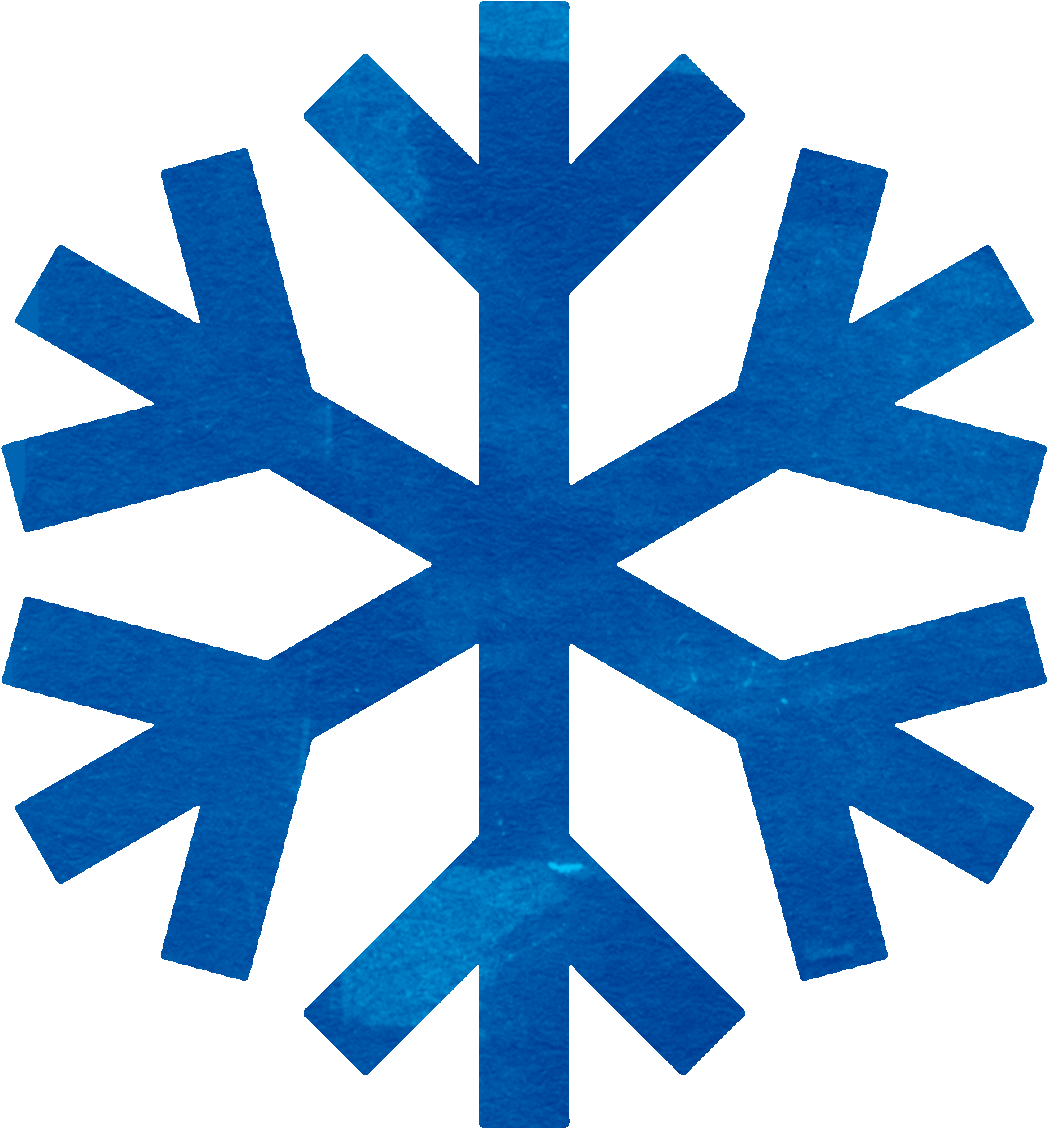} indicates the model is frozen}
    \label{fig:our_system}
\end{figure*}

To overcome this, \citet{mokady_clipcap_2021} train a network to transform the image representation into the domain of the language representation.
They do so by learning the transformation that maps the image representation into a representation that generates an image caption when passed to the LLM.
\citet{tsimpoukelli_multimodal_2021} append a question when passing this transformed image representation to the LLM, for VQA.
This approach is visualized in Figure~\ref{fig:embed-vqa}.

Another approach to VQA is to caption the images, and pass these captions to the LLM as natural language alongside a question \citep{yang_empirical_2022, guo_images_2023, hu_promptcap_2022, tiong_plug-and-play_2022}.
This approach is visualized in Figure~\ref{fig:caption-vqa}

In few-shot VQA, in-context examples are also passed to the LLM in the prompt. 
These are samples of desired behaviour expressed as inputs with correct responses. 
Including in-context examples has been empirically shown to improve few-shot VQA performance \citep{liu-2022-makes, alayrac_flamingo_2022}.

Recent VQA architectures generally employ a variation of either the embedding-based or caption-based approaches. 
However, a comparison of these two approaches has been previously overlooked.
Our contribution is a controlled and focused comparison of these two approaches.
Additionally, we conduct an analysis to determine how the method used to select in-context examples affects the prediction behaviour of the approaches.

\section{Background}

LLMs are transformer-based models shown to be capable of performing text-to-text tasks without task-specific training \citep{chung_scaling_2022}.
This includes extractive question--answering, where the model is prompted to answer a question based on provided context.

In vision--language research, VQA is a fundamental task.
One approach is to pass visual information to an LLM in the form of image captions \citep{yang_empirical_2022, guo_images_2023, hu_promptcap_2022, tiong_plug-and-play_2022}.
The VQA task is then recast as an extractive QA task.

Meanwhile in the field of computer vision, \citet{dosovitskiy_image_2021} present the vision transformer (ViT).
\citet{radford_learning_2021} present a training objective for the ViT, minimizing a contrastive loss between encoded images and captions.
They distribute CLIP models capable of transforming images into generic feature representations.

The primary challenge of integrating LLMs with ViT image encoders is to align features in the image and text space.
\citet{tsimpoukelli_multimodal_2021} fine-tune an image encoder by passing its output into a frozen LLM. A frozen model has weights that are not updated during training. They update the weights of the image encoder using image--caption training data.
They evaluate on the task of VQA by appending embedded questions to the image embeddings and passing to an LLM.

\citet{merullo_linearly_2023} achieve comparable results using only a linear mapping between CLIP \citep{radford_learning_2021} image and GPT-J \citep{gpt-j} text space.
\citet{yi-lin_sung_vl-adapter_2022} and \citet{eichenberg_magma_2022} add a small number of trainable parameters to the LLM, called adapters.
These adapters are trainable MLP networks inserted into the LLM transformer blocks.

\citet{alayrac_flamingo_2022} add masked cross-attention layers in the LLM to ensure each question only attends to the last set of image embeddings.
\citet{li_blip-2_2023} propose a system that includes the prompt question in a querying transformer network.
Both of these systems require vast troves of training data.

The above systems begin by learning an image captioning task.
It is therefore a natural baseline to compare their embedding-based VQA approaches with that of using image captions generated by the same system.
However, none of the aforementioned systems report such results.

\section{Two approaches to VQA}\label{sec:approaches}

We will now outline our embedding-based and caption-based approaches to VQA.
We keep our experiment controlled by using the same image feature representation in both approaches.
The only difference will be whether this visual feature representation is first converted into an image caption, or whether it is passed directly to the LLM.

Both setups use a single trained model. This model is a `mapping network', trained to bridge an image encoder and an LLM. The image encoder and the LLM will be frozen, and the parameters of the mapping network will be learnt by maximising the likelihood of target caption token sequences $\boldsymbol{y}$, given an input image.
Since the image encoder is frozen, this likelihood is equivalent to the likelihood of the target given the image encoder feature embedding~$\boldsymbol{x}$.
This objective is given in Equation~\ref{eqa:objective}, expanded for an autoregressive language model.

\begin{equation}\label{eqa:objective}
    \log{p(\boldsymbol{y}|\boldsymbol{x})} = \sum_{l} \log{p(\boldsymbol{y}_l | \boldsymbol{x}, \boldsymbol{y}_{1:l-1})}
\end{equation}

\begin{enumerate}
    \item \textbf{Image encoder (image to feature embedding)} An image encoder encodes the image pixel value matrix $\boldsymbol{i} \in \mathbb{R}^{W\times H\times 3}$ into a $v$-dimensional vector denoted by $\boldsymbol{x}$. 
    \item \textbf{Mapping network (feature embedding to LLM image representation)} A mapping network will map this feature embedding into $k$ $D$-dimensional embeddings.
    Each of these embeddings is of the same dimensionality as a single embedded textual token.
    \item \textbf{LLM (image representation to text output)} These embeddings are passed to the LLM, generating an output token sequence $\boldsymbol{y}$ which is compared against a groundtruth caption.
\end{enumerate}

Our two approaches to VQA are as follows.

\paragraph{Embedding-based}

The tokenizer and embedder of the LLM transform a question into $t$ embeddings.
An image is converted into $k$ LLM embeddings of the same dimensionality using the image encoder and mapping network.
The mapped image embeddings and question embeddings are concatenated, and together passed to the LLM.

\paragraph{Caption-based}
For caption-based VQA, the LLM image embeddings are passed alone to the LLM to generate an image caption.
This textual image caption is then concatenated with the textual question, and this prompt is passed to the LLM.

\section{Experimental Setup}

\paragraph{Models}

We use Flan-T5 XL \citep{chung_scaling_2022}, a 3B parameter autoregressive encoder--decoder LLM.
We selected this model because it has empirically been shown to perform well with in-context learning.
To encode the images, we use a pre-trained CLIP ViT-G model \citep{radford_learning_2021, ilharco_gabriel_2021_5143773, schuhmann2022laionb}.
We bridge these two components with a single hidden layer mapping network with 1280-D input, a hidden layer of size $20\times2048/2$, and 20 2048-D output embeddings.

\paragraph{Datasets}

We use the conceptual captions dataset \citep{sharma_conceptual_2018} as training data.
From the 3M+ image URLs provided in this dataset, we found 2.7M to be active.

\paragraph{Training}

We minimize cross-entropy loss during training.
We backpropagate losses through the frozen LLM, updating the mapping network with the AdamW optimizer \citep{kingma_adam_2014, loshchilov_decoupled_2017}.
For the optimizer, we use $\beta = (0.9, 0.98)$ and weight decay $=0.01$.
The learning rate grows linearly to 2e-4 for 375 warm-up steps and then decays linearly to zero for two epochs with a batch size of 32.

\paragraph{In-context example selection}\label{sec:in-context-example-selection}

Including samples of desired behaviour, expressed as input with correct responses in the prompt, has been shown to improve performance \citep{yang_empirical_2022}.
This is called in-context learning.
In particular, `shots' refers to the number of samples included.

To find the most similar in-context examples, we extract CLIP embeddings of the train and validation VQAv2 images and questions.
We will compare selecting up to four in-context examples randomly (R), using only question similarity (Q) and joint (50-50) question and image similarity (Q+I).
We use the inner product as our similarity metric, and pre-compute in-context examples efficiently using a FAISS index \citep{johnson2019billion}.

\paragraph{Metrics}

The system is validated few-shot on the VQA v2 benchmark \citep{agrawal_vqa_2015, zhang_yin_2015, goyal_making_2016}, which includes 214K image--question--answer validation triplets.

We use the official evaluation metric, which calculates the number of exact matches between the predicted answer and the 10 human-annotated answers. 
The predicted answer scores if it matches exactly at least three human annotated answers.
We follow recent approaches and evaluate in the open-ended generation setting \citep{tsimpoukelli_multimodal_2021, alayrac_flamingo_2022, li_blip-2_2023}.

Our setup is implemented using the framework of \citet{lin-byrne-2022-retrieval}.

We run each experiment once with a single random seed.
We test the differences between our systems, competitors and baselines for significance with two-tailed paired permutation tests, approximated by $R=10,000$, with threshold $\alpha=0.01$.

\paragraph{Systems, competitors and baselines}.

We will compare results from our two approaches to VQA, \textit{embedding-based} and \textit{caption-based}, with three systems of similar level of data and computation: \textit{Frozen} \citep{tsimpoukelli_multimodal_2021}, \textit{Linear mapping} \citep{merullo_linearly_2023} and \textit{MAGMA} \citep{eichenberg_magma_2022}.
A comparison with these competitors is a sanity check to validate the suitability of our experimental setup.

\section{Results}

Table~\ref{tab:results} gives results on the VQAv2 benchmark\footnote{ 
In 0-shot VQA, no in-context examples are included; so the random (R), question similarity (Q) and question + image similarity (Q+I) in-context example selection variations of our approaches are identical.
We found all other differences to be significant (p$<$0.01).
}.
We can see that our systems pass the sanity check.
Our best results for each of our two approaches to few-shot VQA (lines 6, 9) outperform all three competitor systems (lines 1-3) for all number of shots.

\begin{table}
\footnotesize
\centering
\resizebox{\linewidth}{!}{
\begin{tabular}{rrrrrr}
 &  \multicolumn{4}{c}{\# shots} \\ 
&  0 & 1 & 2 & 4 &  \\ \cmidrule{1-5}
 \textit{Frozen} & 29.5 & 35.7 & - & 38.2 & (1)  \\
 \textit{Linear mapping} & 33.3 & 39.9 & 40.8 & 40.3 & (2) \\
  \textit{MAGMA} & 36.9 & 42.2 & 43.8 & 45.4 & (3)  \\ 

\cmidrule{1-5}
 \textit{(R)} & \textbf{45.9} & 45.3 & 45.1 & 45.3 & (4) \\
 \textit{Caption-based} \quad \textit{(Q)} & \textbf{45.9} & 47.6 & 48.0 & 48.5 & (5) \\
  \textit{(Q+I)} & \textbf{45.9} & 50.0 & 50.3 & 50.8 & (6) \\
\cmidrule{1-5}
  \textit{(R)} & 41.4 & 44.5 & 43.6 & 42.5 & (7) \\
  \textit{Embedding-based} \textit{(Q)} & 41.4 & 40.6 & 45.7 & 47.3 & (8) \\
  \textit{(Q+I)} & 41.4 & \textbf{50.5} & \textbf{51.8} & \textbf{52.4} & (9) \\
  
\cmidrule{1-5}

\end{tabular}
}
\caption{Few-shot VQAv2 results
}
\label{tab:results}
\end{table}

Our results (lines 4-6 vs. lines 7-9) show that 0-shot caption-based VQA is significantly better (+4.5\%) than 0-shot embedding-based VQA. 
This is a surprising result, because the same visual representation is used in both approaches.
The only difference is that in caption-based VQA, the embeddings are first passed to the LLM alone to generate a caption, before being concatenated with the question.
This result is a clear indication that the caption-based approach to VQA is an important comparison that should be made for systems that begin by learning the mapping between image and text space with an image captioning task.

Our results (line 9) show a large improvement (+9.1\%) between zero and 1-shot embedding-based VQA: where a single in-context example is included.
The in-context example appears to demonstrate the task in a way that has a substantial improvement to the VQA capability of the LLM.
This is evidence of the in-context learning approach working well, enabling our system to adapt to the task of VQA without task-specific training.

Furthermore, all our results from selecting in-context examples based on question (lines 5, 8) or question + image (lines 6, 9) similarity are significantly better than selecting in-context examples randomly (lines 4, 7).
Giving relevant (similar to the test question/image) examples is better than giving random examples of the task.
This is additional evidence that the in-context learning approach is performing as intended.

In the 1-shot setting, the embedding-based (line 9) approach is better than the caption-based approach (line 6, +0.5\%). 
This trend continues; the results improve by small margins, and the embedding-based approach is slightly better than using image captions, for 2- and 4-shots.
The results plateau with increasing number of in-context examples (shots), which aligns with prior works \citep{tsimpoukelli_multimodal_2021, alayrac_flamingo_2022}.

We explored a range of comparisons between the caption-based and embedding-based approaches to compare their prediction behaviour. 
One way we did this was by varying how the in-context examples are selected.
Our main finding was that the two approaches vary in their responsiveness to the in-context examples.

The most striking result is that of embedding-based VQA, when in-context examples are selected based on top question similarity (line 8).
Here, two shots/in-context examples are required to demonstrate the task suitably; at this point, the improvement is still smaller than before (+4.3\%).
We find that the caption-based approach beats embedding-based VQA (line 5 vs. line 8) for all number of shots when the in-context example is selected based only on question similarity.

We additionally explored the accuracy of our systems in various specific question categories, such as those that ask for colour or counting information. 
Our analysis reveals that 4-shot VQA with the embedding-based approach correctly answers more questions requiring small visual details compared to the caption-based approach (e.g. colour: +4.2\%, counting: +2.7\%), while compensating in straightforward questions (e.g. yes/no: -0.5\%). 
For complex `why' questions or location `where' questions, where our systems are weak, both demonstrate similar prediction behaviour.

\section{Limitations}\label{sec:discussion}

Our findings are applicable for the Flan-T5 XL LLM, which is not as large as other closed-source LLMs. 
Our tale may have been different for these models.
In addition, we learn the mapping between image and text space using a non-linear network, rather than other methods such as that presented by \citet{li_blip-2_2023}.
Instead we have presented a controlled and focused setup, with the intention of initiating an investigation of the caption-based and embedding-based approaches to few-shot VQA.

\section{Conclusion}

In this work, we have presented a focused comparison of two approaches to few-shot VQA.
A comparison of these two approaches has been overlooked in previous work.
We learn a non-linear mapping between the CLIP and Flan-T5 XL embedding space, which enables us to generate both image captions and an embedding-based LLM image representation from the same system.
Our results indicate that, using this non-linear mapping and the relatively small frozen LLM, connecting visual embeddings directly to the LLM embedding space does not necessarily improve performance compared to using image captions. 
We find that the selection of relevant in-context examples is far more important.
Overall in this work, we hope to have shown that reporting results with system-generated captions is an important comparison for multimodal systems that pass embedding-based visual representations to LLMs.

\bibliography{acl2023}
\bibliographystyle{acl_natbib}

\end{document}